\documentclass[runningheads]{llncs}
\usepackage{graphicx}
\usepackage{colortbl}
\usepackage{comment}
\usepackage{amsmath,amssymb} 

\graphicspath{ {./images/} }
\usepackage{makecell}
\usepackage{multirow}
\usepackage{color, colortbl}
\usepackage{soul}
\usepackage{stackengine}
\usepackage{csquotes}
\usepackage{amsmath,bm, bbm}
\usepackage[ruled, lined, linesnumbered, longend]{algorithm2e}
\SetAlFnt{\scriptsize}
\SetAlCapFnt{\small}
\SetAlCapNameFnt{\small}
\let\oldnl\nl
\newcommand{\nonl}{\renewcommand{\nl}{\let\nl\oldnl}}
\usepackage{amsmath}
\usepackage{subfigure}
\usepackage{caption}
\usepackage{footmisc}
\usepackage[normalem]{ulem}
\usepackage{booktabs,array,siunitx}
\definecolor{Gray}{gray}{0.95}
\usepackage{cuted}
\usepackage{capt-of}
\newcommand\xrowht[2][0]{\addstackgap[.5\dimexpr#2\relax]{\vphantom{#1}}}

\newcommand{\eg}{\emph{e.g.}}
\newcommand{\ie}{\emph{i.e.}}
\newcommand{\etal}{\emph{et al.}}
\newcommand{\etc}{\emph{etc.}}

\usepackage[accsupp]{axessibility}  

\usepackage{xcolor}
\usepackage{hyperref}
\hypersetup{colorlinks=true,linkcolor=red,citecolor=green,urlcolor=blue}
\begin{document}
\pagestyle{headings}
\mainmatter

\def\ACCV22SubNumber{478}  

\title{PU-Transformer: Point Cloud Upsampling Transformer} 

\titlerunning{PU-Transformer: Point Cloud Upsampling Transformer}
%
\author{Shi Qiu\inst{1,2} \and
Saeed Anwar\inst{1,2} \and
Nick Barnes\inst{1}}
\authorrunning{S. Qiu et al.}
%
\institute{Australian National University \and
Data61-CSIRO, Australia\\
\email{\{shi.qiu, saeed.anwar, nick.barnes\}@anu.edu.au}}
\maketitle

\begin{abstract}
Given the rapid development of 3D scanners, point clouds are becoming popular in AI-driven machines. However, point cloud data is inherently sparse and irregular, causing significant difficulties for machine perception. In this work, we focus on the point cloud upsampling task that intends to generate dense high-fidelity point clouds from sparse input data. Specifically, to activate the transformer's strong capability in representing features, we develop a new variant of a multi-head self-attention structure to enhance both point-wise and channel-wise relations of the feature map. In addition, we leverage a positional fusion block to comprehensively capture the local context of point cloud data, providing more position-related information about the scattered points. As the first transformer model introduced for point cloud upsampling, we demonstrate the outstanding performance of our approach by comparing with the state-of-the-art CNN-based methods on different benchmarks quantitatively and qualitatively.
\end{abstract}

\section{Introduction}
\label{sec:intro}
3D computer vision has been attracting a wide range of interest from academia and industry since it shows great potential in many fast-developing AI-related applications such as robotics, autonomous driving, augmented reality, \etc ~As a basic representation of 3D data, point clouds can be easily captured by 3D sensors~\cite{endres20133,jaboyedoff2012use}, incorporating the rich context of real-world surroundings.

Unlike well-structured 2D images, point cloud data has inherent properties of \emph{irregularity} and \emph{sparsity}, posing enormous challenges for high-level vision tasks such as point cloud classification~\cite{qi2017pointnet,wang2019dynamic,qiu2021geometric}, segmentation~\cite{qi2017pointnet++,qiu2021dense,hu2020randla}, and object detection~\cite{qi2019deep,qi2020imvotenet,qiu2021investigating}. For instance, Uy~\etal~\cite{uy2019revisiting} fail to classify the real-world point clouds while they apply a pre-trained model of synthetic data; and recent 3D segmentation and detection networks~\cite{hu2020randla,qiu2021semantic,Park_2021_ICCV} achieve \emph{worse} results on the distant/smaller objects (\eg, bicycles, traffic-signs) than the closer/larger objects (\eg, vehicles, buildings). If we mitigate point cloud data's \emph{irregularity} and \emph{sparsity}, further improvements in visual analysis can be obtained (as verified in~\cite{ye2021meta}). Thus, point cloud upsampling deserves a deeper investigation.

\begin{figure*}
\begin{center}
\includegraphics[width=0.9\textwidth]{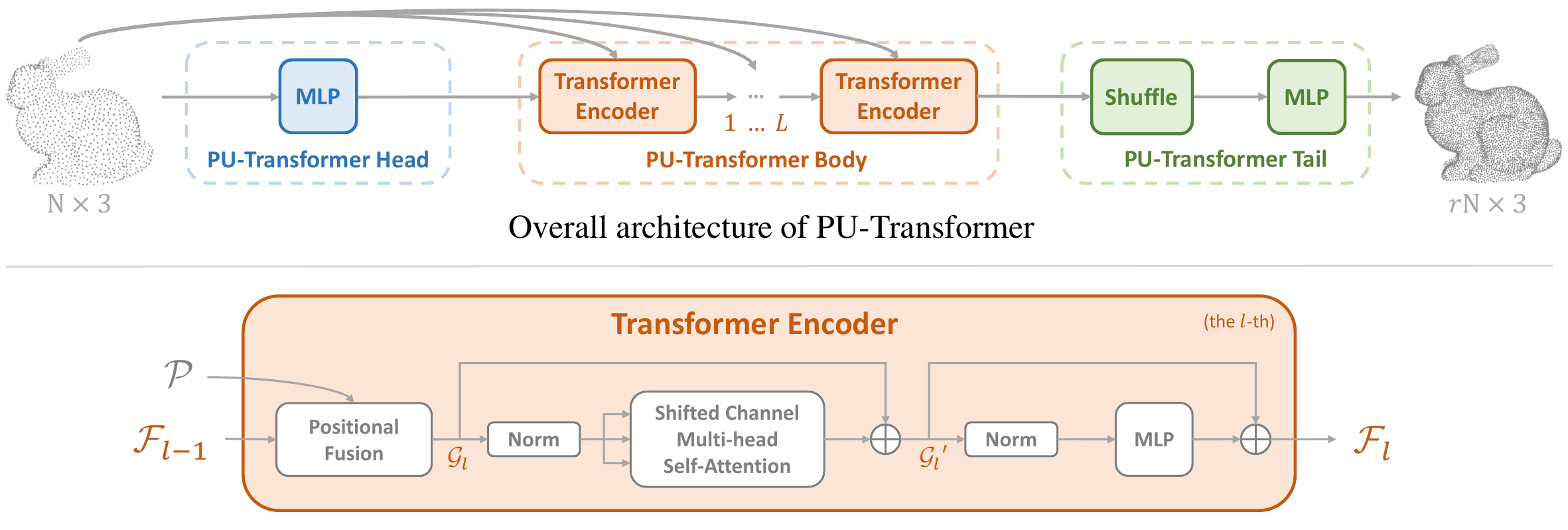}
\end{center}

\captionsetup{font=small}
\caption{The details of PU-Transformer. The upper chart shows the overall architecture of the PU-Transformer model containing three main parts: the PU-Transformer head (Sec.~\ref{sec:head}), body (Sec.~\ref{sec:pu_body}), and tail (Sec.~\ref{sec:tail}). The PU-Transformer body includes a cascaded set of Transformer Encoders (\eg, $L$ in total), serving as the core component of the whole model. Particularly, the detailed structure of each Transformer Encoder is shown in the lower chart, where all annotations are consistent with Line 3-5 in Alg.~\ref{alg:put}.}
\label{fig:net}

\end{figure*}

As a basic 3D low-level vision task, point cloud upsampling aims to generate dense point clouds from sparse input, where the generated data should recover the fine-grained structures at a higher resolution. Moreover, the upsampled points are expected to lie on the underlying surfaces in a uniform distribution, benefiting downstream tasks for both 3D visual analysis~\cite{liu2019relation,qiu2021pnp} and graphic modeling~\cite{mitra2003estimating,mitra2004registration}. Following the success of Convolution Neural Networks (CNNs) in image super-resolution~\cite{dong2015image,kim2016accurate,anwar2020deep} and Multi-Layer-Perceptrons (MLPs) in point cloud analysis~\cite{qi2017pointnet,qi2017pointnet++}, previous methods tended to upsample point clouds via complex network designs (\eg, Graph Convolutional Network~\cite{qian2021pu}, Generative Adversarial Network~\cite{li2019pu}) and dedicated upsampling strategies (\eg, progressive training~\cite{yifan2019patch}, coarse-to-fine reconstruction~\cite{liu2020spu}, disentangled refinement~\cite{li2021point}). As far as we are concerned, these methods share a key to point cloud upsampling: learning the representative features of given points to estimate the distribution of new points. Considering that regular MLPs have limited-expression and generalization capability, we need a more powerful tool to extract fine-grained point feature representations for high-fidelity upsampling. To this end, we introduce a succinct transformer model, PU-Transformer, to effectively upsample point clouds following a simple pipeline as illustrated in Fig.~\ref{fig:net}. The main reasons for adopting transformers to point cloud upsampling are as follows:

\emph{Plausibility in theory.} As the core operation of transformers, self-attention~\cite{vaswani2017attention} is a set operator~\cite{zhao2021point} calculating long-range dependencies between elements regardless of data order. On this front, self-attention can easily estimate the point-wise dependencies without any concern for the inherent \emph{unorderedness}. However, to comprehensively represent point cloud features, channel-wise information is also shown to be a crucial factor in attention mechanisms~\cite{qiu2021geometric,qiu2021investigating}. Moreover, such channel-wise information enables an efficient upsampling via a simple periodic shuffling~\cite{shi2016real} operated on the channels of point features, saving complex designs~\cite{liu2020spu,li2019pu,li2021point,yifan2019patch} for upsampling strategy. Given these facts, we propose a Shifted Channel Multi-head Self-Attention (SC-MSA) block, which strengthens the point-wise relations in a multi-head form and enhances the channel-wise connections by introducing the overlapping channels between consecutive heads.

\emph{Feasibility in practice.} Since the transformer model was originally invented for natural language processing; its usage has been widely recognized in high-level visual applications for 2D images~\cite{carion2020end,dosovitskiy2020image,liu2021swin}. More recently, Chen~\etal~\cite{chen2021pre} introduced a pre-trained transformer model achieving excellent performance on image super-resolution and denoising. Inspired by the transformer's effectiveness for image-related low-level vision tasks, we attempt to create a transformer-based model for point cloud upsampling. Given the mentioned differences between 2D images and 3D point clouds, we introduce the Positional Fusion block as a replacement for positional encoding in conventional transformers: on the one hand, local information is aggregated from both the \emph{geometric} and \emph{feature} context of the points, implying their 3D positional relations; on the other hand, such \emph{local} information can serve as complementary to subsequent self-attention operations, where the point-wise dependencies are calculated from a \emph{global} perspective. 

\emph{Adaptability in various applications.} Transformer-based models are considered as a luxury tool in computer vision due to the huge consumption of data, hardware, and computational resources. However, our PU-Transformer can be easily trained with a \emph{single} GPU in a few hours, retaining a similar model complexity to regular CNN-based point cloud upsampling networks~\cite{yu2018pu,yifan2019patch,li2021point}. Moreover, following a patch-based pipeline~\cite{yifan2019patch}, the trained PU-Transformer model can effectively and flexibly upsample different types of point cloud data, including but not limited to regular object instances or large-scale LiDAR scenes (as shown in Fig.~\ref{fig:vis},~\ref{fig:vis_scan} and~\ref{fig:scan}). Starting with the upsampling task in low-level vision, we expect our approach to transformers will be affordable in terms of resource consumption for more point cloud applications. Our main contributions are:
\begin{itemize}
\item
To the best of our knowledge, we are the first to introduce a transformer-based model\footnote{The project page is: \url{https://github.com/ShiQiu0419/PU-Transformer}.} for point cloud upsampling.

\item We quantitatively validate the effectiveness of the PU-Transformer by significantly outperforming the results of state-of-the-art point cloud upsampling networks on two benchmarks using three metrics.

\item The upsampled visualizations demonstrate the superiority of PU-Transformer for diverse point clouds.
\end{itemize}
\section{Related Work}
\label{sec:work}
\noindent \textbf{Point Cloud Networks:}
In early research, the projection-based methods~\cite{su2015multi,lawin2017deep} used to project 3D point clouds into multi-view 2D images, apply regular 2D convolutions and fuse the extracted information for 3D analysis. Alternatively, discretization-based approaches~\cite{guo2020deep} tended to convert the point clouds to voxels~\cite{huang2016point} or lattices~\cite{su2018splatnet}, and then process them using 3D convolutions or sparse tensor convolutions~\cite{choy20194d}. To avoid context loss and complex steps during data conversion, the point-based networks~\cite{qi2017pointnet,qi2017pointnet++,wang2019dynamic}  directly process point cloud data via MLP-based operations. Although current mainstream approaches in point cloud upsampling prefer utilizing MLP-related modules, in this paper, we focus on an advanced transformer structure~\cite{vaswani2017attention} in order to further enhance the point-wise dependencies between known points and benefit the generation of new points.

\noindent \textbf{Point Cloud Upsampling:}
Despite the fact that current point cloud research in low-level vision~\cite{yu2018pu,yuan2018pcn} is less active than that in high-level analysis~\cite{qi2017pointnet,hu2020randla,qi2019deep}, there exists many outstanding works that have contributed significant developments to the point cloud upsampling task. To be specific, PU-Net~\cite{yu2018pu} is a pioneering work that introduced CNNs to point cloud upsampling based on a PointNet++~\cite{qi2017pointnet++} backbone. Later, MPU~\cite{yifan2019patch} proposed a patch-based upsampling pipeline, which can flexibly upsample the point cloud patches with rich local details. In addition, PU-GAN~\cite{li2019pu} adopted the architecture of Generative Adversarial Networks~\cite{goodfellow2014generative} for the generation problem of high-resolution point clouds, while PUGeo-Net~\cite{qian2020pugeo} indicated a promising combination of discrete differential geometry and deep learning. More recently, Dis-PU~\cite{li2021point} applies disentangled refinement units to gradually generate the high-quality point clouds from coarse ones, and PU-GCN~\cite{qian2021pu} achieves good upsampling performance by using graph-based network constructions~\cite{wang2019dynamic}. Moreover, there are some papers exploring \emph{flexible-scale} point cloud upsampling via meta-learning~\cite{ye2021meta}, self-supervised learning~\cite{zhao2021sspu}, decoupling ratio with network architecture~\cite{luo2021pu}, or interpolation~\cite{qian2021deep}, \etc~As the first work leveraging transformers for point cloud upsampling, we focus on the effectiveness of PU-Transformer in performing the fundamental \emph{fixed-scale} upsampling task, and expect to inspire more future work in relevant topics. 

\noindent \textbf{Transformers in Vision:}
With the capacity in parallel processing as well as the scalability to deep networks and large datasets~\cite{khan2021transformers}, more visual transformers have achieved excellent performance on image-related tasks including either low-level~\cite{yang2020learning,chen2021pre} or high-level analysis~\cite{dosovitskiy2020image,liu2021swin,carion2020end,zhu2020deformable}. Due to the inherent gaps between 3D and 2D data, researchers introduce the variants of transformer for point cloud analysis~\cite{yew2022regtr,fan2021point,yu2022point,fan2022point,yu2021pointr}, using vector-attention~\cite{zhao2021point}, offset-attention~\cite{guo2021pct}, and grid-rasterization~\cite{mazur2021cloud}, \etc ~However, since these transformers still operate on an overall classical PointNet~\cite{qi2017pointnet} or PointNet++ architecture~\cite{qi2017pointnet++}, the improvement is relatively limited while the computational cost is too expensive for most researchers to re-implement. To simplify the model's complexity and boost its adaptability in point cloud upsampling research, we only utilize the general structure of transformer encoder~\cite{dosovitskiy2020image} to form the body of our PU-Transformer.

\section{Methodology}
\label{sec:metho}
\subsection{Overview}
As shown in Fig.~\ref{fig:net}, given a sparse point cloud $\mathcal{P}\in\mathbb{R}^{N\times3}$, our proposed PU-Transformer can generate a dense point cloud $\mathcal{S}\in\mathbb{R}^{rN\times3}$, where $r$ denotes the upsampling scale. Firstly, the PU-Transformer head extracts a preliminary feature map from the input. Then, based on the extracted feature map and the inherent 3D coordinates, the PU-Transformer body gradually encodes a more comprehensive feature map via the cascaded Transformer Encoders. Finally, in the PU-Transformer tail, we use the shuffle operation~\cite{shi2016real} to form a dense feature map and reconstruct the 3D coordinates of $\mathcal{S}$ via an MLP.

In Alg.~\ref{alg:put}, we present the basic operations that are employed to build our PU-Transformer. As well as the operations (\enquote{MLP}~\cite{qi2017pointnet}, \enquote{Norm}~\cite{ba2016layer}, \enquote{Shuffle}~\cite{shi2016real}) that have been widely used in image and point cloud analysis, we propose two novel blocks targeting a transformer-based point cloud upsampling model \ie, the Positional Fusion block (\enquote{\textbf{PosFus}} in Alg.~\ref{alg:put}), and the Shifted-Channel Multi-head Self-Attention block (\enquote{\textbf{SC-MSA}} in Alg.~\ref{alg:put}). In the rest of this section, we introduce these two blocks in detail. Moreover, for a compact description, we only consider the case of an \emph{arbitrary} Transformer Encoder; thus, in the following, we discard the subscripts that are annotated in Alg.~\ref{alg:put} denoting a Transformer Encoder's specific index in the PU-Transformer body.
\noindent
\begin{figure}[t]
\begin{minipage}[t]{0.65\textwidth}
    \vspace{0pt}
    \footnotesize
    \begin{algorithm}[H]
    \caption{PU-Transformer Pipeline}\label{alg:put}
    \nonl\textbf{input:} a sparse point cloud $\mathcal{P}\in\mathbb{R}^{N\times3}$\\
    \nonl\textbf{output:} a dense point cloud $\mathcal{S}\in\mathbb{R}^{rN\times3}$\\
    {\nonl{\fontfamily{qcr}\selectfont\textcolor{gray}{\# PU-Transformer Head}}}\\
    $\mathcal{F}_0$ = MLP($\mathcal{P}$)\\
    {\nonl{\fontfamily{qcr}\selectfont\textcolor{gray}{\# PU-Transformer Body}}}\\
    \For{each Transformer Encoder}{
        {\nonl{\fontfamily{qcr}\selectfont\textcolor{gray}{\# $l = 1\;...\;L$}}}\\
        {\nonl{\fontfamily{qcr}\selectfont\textcolor{gray}{\# the $l$-th Transformer Encoder}}}\\
        $\mathcal{G}_{l}$ = \textbf{PosFus}($\mathcal{P}$, $\mathcal{F}_{l-1}$)\;
        ${\mathcal{G}_{l}}^\prime$ = \textbf{SC-MSA}\big(Norm($\mathcal{G}_{l}$)\big) + $\mathcal{G}_{l}$\;
        $\mathcal{F}_{l}$ = MLP\big(Norm(${\mathcal{G}_{l}}^\prime$)\big) + ${\mathcal{G}_{l}}^\prime$\;
    }
    {\nonl{\fontfamily{qcr}\selectfont\textcolor{gray}{\# PU-Transformer Tail}}}\\
    $\mathcal{S}$ = MLP\big(Shuffle($\mathcal{F}_L$)\big)
    \end{algorithm}
\end{minipage}
\end{figure}
\subsection{Positional Fusion}
\label{sec:posfus}
Usually, a point cloud consisting of $N$ points has two main types of context: the 3D coordinates $\mathcal{P}\in\mathbb{R}^{N\times3}$ that are explicitly sampled from synthetic meshes or captured by real-world scanners, showing the original geometric distribution of the points in 3D space; and the feature context, $\mathcal{F}\in\mathbb{R}^{N\times C}$, that is implicitly encoded by convolutional operations in $C$-dimensional embedding space, yielding rich latent clues for visual analysis. Older approaches~\cite{yu2018pu,yifan2019patch,li2019pu} to point cloud upsampling generate a dense point set by heavily exploiting the encoded features $\mathcal{F}$, while recent methods~\cite{qian2020pugeo,qian2021pu} attempt to incorporate more geometric information. As the core module of the PU-Transformer, the proposed Transformer Encoder leverages a Positional Fusion block to encode and combine both the given $\mathcal{P}$ and $\mathcal{F}$\footnote{equivalent to \enquote{$\mathcal{F}_{l-1}$} in Alg.~\ref{alg:put}} of a point cloud, following the local geometric relations between the scattered points.

Based on the metric of \emph{3D-Euclidean distance}, we can search for neighbors ${\forall}p_j\in Ni(p_i)$ for each point $p_i\in\mathbb{R}^{3}$ in the given point cloud $\mathcal{P}$, using the k-nearest-neighbors (knn) algorithm~\cite{wang2019dynamic}. Coupled with a grouping operation, we thus obtain a matrix $\mathcal{P}_j\in\mathbb{R}^{N\times k \times3}$, denoting the 3D coordinates of the neighbors for all points. Accordingly, the relative positions between each point and its neighbors can be formulated as:
\begin{equation}
\label{equ:1}
    \Delta\mathcal{P} = \mathcal{P}_j - \mathcal{P}, \quad \Delta\mathcal{P}\in\mathbb{R}^{N\times k \times3};
\end{equation}
where $k$ is the number of neighbors. In addition to the neighbors' relative positions showing each point's local detail, we also append the centroids' positions in 3D space, indicating the global distribution for all points. By duplicating $\mathcal{P}$ in a dimension expanded $k$ times, we concatenate the local \emph{geometric} context:
\begin{equation}
\label{equ:2}
    \mathcal{G}_{geo} = \mathrm{concat}\big[\underset{k}{\mathrm{dup}}(\mathcal{P}); \Delta\mathcal{P}\big]\in\mathbb{R}^{N\times k \times6}.
\end{equation}

Further, for the feature matrix $\mathcal{F}_j\in\mathbb{R}^{N\times k \times C}$ of all searched neighbors, we conduct similar operations (Eq.~\ref{equ:1} and~\ref{equ:2}) as on the counterpart $\mathcal{P}_j$, computing the relative features as:
\begin{equation}
\label{equ:3}
    \Delta\mathcal{F} = \mathcal{F}_j - \mathcal{F}, \quad \Delta\mathcal{F}\in\mathbb{R}^{N\times k \times C};
\end{equation}
and representing the local \emph{feature} context as:
\begin{equation}
\label{equ:4}
    \mathcal{G}_{feat} = \mathrm{concat}\big[\underset{k}{\mathrm{dup}}(\mathcal{F}); \Delta\mathcal{F}\big]\in\mathbb{R}^{N\times k \times 2C}.
\end{equation}

After the local \emph{geometric} context $\mathcal{G}_{geo}$ and local \emph{feature} context $\mathcal{G}_{feat}$ are constructed, we then fuse them for a comprehensive point feature representation. Specifically, $\mathcal{G}_{geo}$ and $\mathcal{G}_{feat}$ are encoded via two MLPs, $\bm{\mathcal{M}}_{\Phi}$ and $\bm{\mathcal{M}}_{\Theta}$, respectively; further, we comprehensively aggregate the local information, $\mathcal{G}\in\mathbb{R}^{N\times C^\prime}$\footnote{equivalent to \enquote{$\mathcal{G}_{l}$} in Alg.~\ref{alg:put}}, using a concatenation between the encoded two types of local context, followed by a max-pooling function operating over the neighborhoods. The above operations can be summarized as:
\begin{equation}
\label{equ:g}
    \mathcal{G} = \underset{k}{\mathrm{max}}\Big(\mathrm{concat}\big[\bm{\mathcal{M}}_{\Phi}(\mathcal{G}_{geo}); \bm{\mathcal{M}}_{\Theta}(\mathcal{G}_{feat})\big]\Big).
\end{equation}

Unlike the local graphs in DGCNN~\cite{wang2019dynamic} that need to be updated in every encoder based on the \emph{dynamic} relations in embedding space, both of our $\mathcal{G}_{geo}$ and $\mathcal{G}_{feat}$ are constructed (\ie, Eq.~\ref{equ:2} and~\ref{equ:4}) and encoded (\ie, $\bm{\mathcal{M}}_{\Phi}$ and $\bm{\mathcal{M}}_{\Theta}$ in Eq.~\ref{equ:g}) in the same way, following \emph{fixed} 3D geometric relations (\ie, ${\forall}p_j\in Ni(p_i)$ defined upon \emph{3D-Euclidean distance}). The main benefits of our approach can be concluded from two aspects: (i) it is practically efficient since the expensive knn algorithm just needs to be conducted once, while the searching results can be utilized in all Positional Fusion blocks of the PU-Transformer body; and (ii) the local \emph{geometric} and \emph{feature} context are represented in a similar manner following the same metric, contributing to \emph{fairly fusing} the two types of context. A detailed behavior analysis of this block is provided in the supplementary material. 

Overall, the Positional Fusion block can not only encode the positional information about a set of unordered points for the transformer's processing, but also aggregate comprehensive local details for accurate point cloud upsampling.

\subsection{Shifted Channel Multi-head Self-Attention}
\label{sec:body}
Different from previous works that applied complex upsampling strategies (\eg, GAN~\cite{li2019pu}, coarse-to-fine~\cite{liu2020spu}, task-disentangling~\cite{li2021point}) to estimate new points, we prefer generating dense points in a simple way. Particularly, PixelShuffle~\cite{shi2016real} is a periodic shuffling operation that efficiently reforms the \emph{channels} of each point feature to represent new points without introducing additional parameters. However, with regular multi-head self-attention (MSA)~\cite{vaswani2017attention} serving as the main calculation unit in transformers, only \emph{point-wise} dependencies are calculated in each independent head of MSA, lacking integration of \emph{channel-related} information for shuffling-based upsampling. To tackle this issue, we introduce a Shifted Channel Multi-head Self-Attention (SC-MSA) block for the PU-Transformer.

\begin{figure}[t]
\noindent
\begin{minipage}[t]{0.55\textwidth}
  \vspace{0pt}  
  \begin{algorithm}[H]
\caption{Shifted Channel\\Multi-head Self-Attention (\textbf{SC-MSA})}\label{alg:msa}
\nonl\textbf{input:} a point cloud feature map: $\mathcal{I}\in\mathbb{R}^{N\times C^\prime}$\\
\nonl\textbf{output:} the refined feature map: $\mathcal{O}\in\mathbb{R}^{N\times C^\prime}$\\
\nonl\textbf{others:} channel-wise split width: $w$ \\
\nonl\quad \quad \,\,\,\,\,\, channel-wise shift interval: $d$, $d<w$ \\
\nonl\quad \quad \,\,\,\,\,\, the number of heads: $M$ \\
$\mathcal{Q} = \mathrm{Linear}(\mathcal{I})$ \quad {\fontfamily{qcr}\selectfont\textcolor{gray}{\# Query Mat $\mathcal{Q}\in\mathbb{R}^{N\times C^\prime}$}}\\
$\mathcal{K} = \mathrm{Linear}(\mathcal{I})$ \quad {\fontfamily{qcr}\selectfont\textcolor{gray}{\# Key Mat $\mathcal{K}\in\mathbb{R}^{N\times C^\prime}$}}\\
$\mathcal{V} = \mathrm{Linear}(\mathcal{I})$ \quad {\fontfamily{qcr}\selectfont\textcolor{gray}{\# Value Mat $\mathcal{V}\in\mathbb{R}^{N\times C^\prime}$}}\\
\For{$m\in \{1, 2, ..., M\}$}{
    $\mathcal{Q}_m = \mathcal{Q}[\;:\;, (m-1)d:(m-1)d+w]$\;
    $\mathcal{K}_m = \mathcal{K}[\;:\;, (m-1)d:(m-1)d+w]$\;
    $\mathcal{V}_m = \mathcal{V}[\;:\;, (m-1)d:(m-1)d+w]$\;
    $\mathcal{A}_m = \mathrm{softmax} (\mathcal{Q}_m {\mathcal{K}_m}^{T})$\;
    $\mathcal{O}_m = \mathcal{A}_m \mathcal{V}_m$\;
}
\textbf{obtain:} $\{\mathcal{O}_1, \mathcal{O}_2, ..., \mathcal{O}_M\}$\\
$\mathcal{O} = \mathrm{Linear}\Big(\mathrm{concat}\big[\{\mathcal{O}_1, \mathcal{O}_2, ..., \mathcal{O}_M\}\big]\Big)$\\
  \end{algorithm}
\end{minipage}%
\begin{minipage}[t]{0.45\textwidth}
  \vspace{10pt}
  \centering
    \includegraphics[width=\columnwidth]{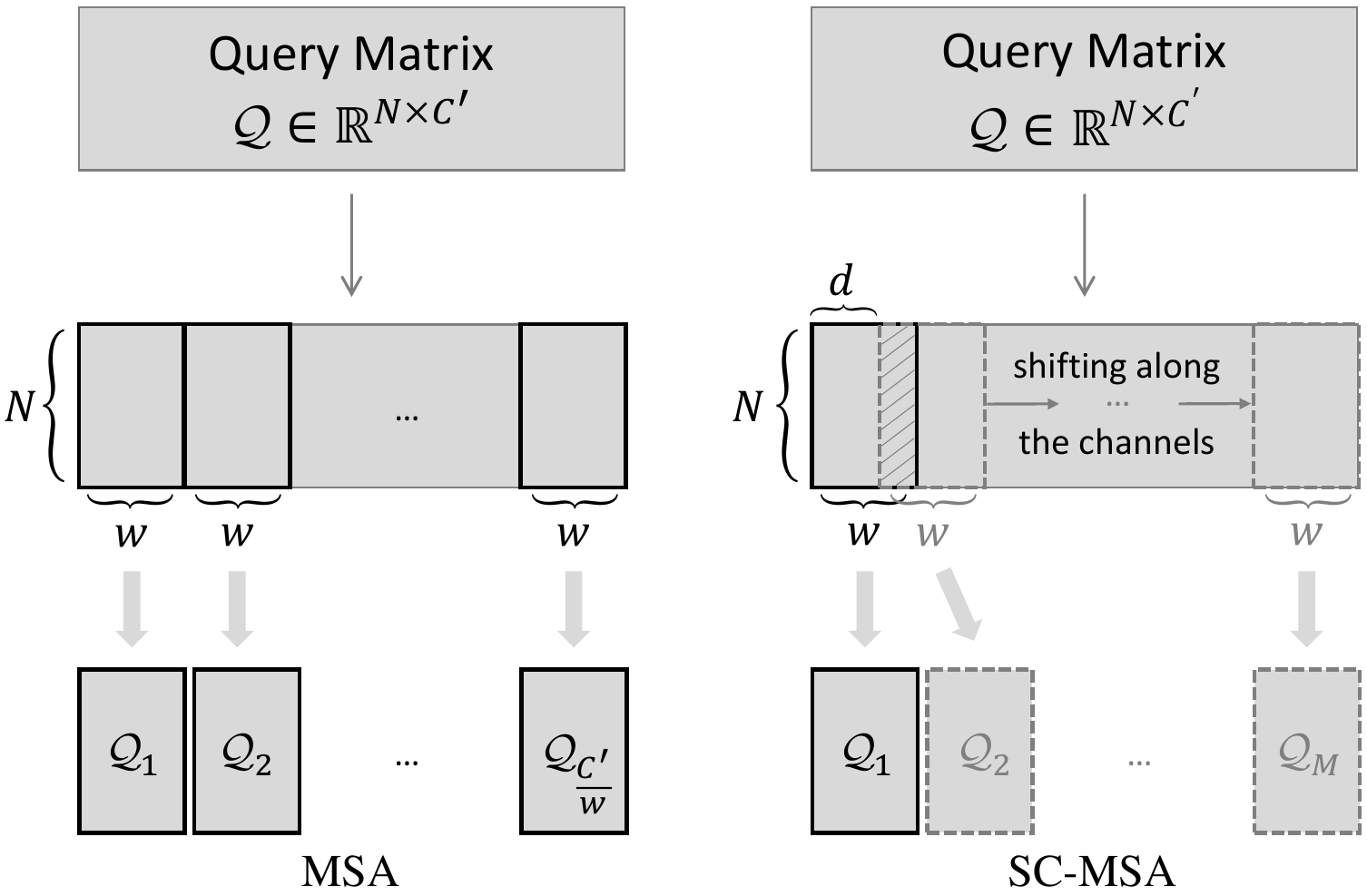}
    \captionof{figure}{\small Examples of how regular MSA~\cite{vaswani2017attention} and our SC-MSA generate the low-dimensional splits of query matrix $\mathcal{Q}$ for multi-head processing (the same procedure applies to $\mathcal{K}$ and $\mathcal{V}$).}
    \label{fig:splits}
\end{minipage}
\end{figure}

As Alg.~\ref{alg:msa} states, at first, we apply linear layers (denoted as \enquote{$\mathrm{Linear}$}, and implement as a $1\times1$ convolution) to encode the query matrix $\mathcal{Q}$, key matrix $\mathcal{K}$, and value matrix $\mathcal{V}$. Then, we generate low-dimensional splits of $\mathcal{Q}_m,\mathcal{K}_m,\mathcal{V}_m$ for each head. Particularly, as shown in Fig.~\ref{fig:splits}, regular MSA generates the \emph{independent} splits for the self-attention calculation in corresponding heads. In contrast, our SC-MSA applies a window (dashed square) shift along the channels to ensure that any two consecutive splits have an overlap of $(w-d)$ channels (slashed area), where $w$ is the channel dimension of each split and $d$ represents the channel-wise shift interval each time. After generating the $\mathcal{Q}_m,\mathcal{K}_m,\mathcal{V}_m$ for each head in the mentioned manner, we employ self-attention (Alg.~\ref{alg:msa} steps 8-9) to estimate the point-wise dependencies as the output $\mathcal{O}_m$ of each head. Considering the fact that any two consecutive heads have part of the input in common (\ie, the overlap channels), thus the connections between the outputs $\{\mathcal{O}_1, \mathcal{O}_2, ..., \mathcal{O}_M\}$ (Alg.~\ref{alg:msa} step 11) of multiple heads are established. There are two major benefits of such connections: (i) it is easier to integrate the information between the \emph{connected} multi-head outputs (Alg.~\ref{alg:msa} step 12), compared to using the \emph{independent} multi-head results of regular MSA; and (ii) as the overlapping context is captured from the channel dimension, our SC-MSA can further enhance the channel-wise relations in the final output $\mathcal{O}$, better fulfilling an efficient and effective shuffling-based upsampling strategy than only using regular MSA's point-wise information. These benefits contribute to a faster training convergence and a better upsampling performance, especially when we deploy fewer Transformer Encoders. More practical evidence is provided in the supplementary material.  

It is worth noting that SC-MSA requires the shift interval to be smaller than the channel-wise width of each split (\ie, $d<w$ as in Alg.~\ref{alg:msa}) for a shared area between any two consecutive splits. Accordingly, the number of heads in our SC-MSA is higher than regular MSA (\ie, $M > C^{\prime}/w$ in Fig.~\ref{fig:splits}). More implementation detail and the choices of parameters are provided in Sec.~\ref{sec:pu_body}.

\section{Implementation}
\label{sec:impl}
\subsection{PU-Transformer Head}
\label{sec:head}
As illustrated in Fig.~\ref{fig:net}, our PU-Transformer model begins with the head to encode a preliminary feature map for the following operations. In practice, we only use a single layer MLP (\ie, a single $1\times1$ convolution, followed by a batch normalization layer~\cite{ioffe2015batch} and a ReLU activation~\cite{nair2010rectified}) as the PU-Transformer head, where the generated feature map size is $N\times16$.

\subsection{PU-Transformer Body}
\label{sec:pu_body}
To balance the model complexity and effectiveness, empirically, we leverage \emph{five} cascaded Transformer Encoders (\ie, $L=5$ in Alg.~\ref{alg:put} and Fig.~\ref{fig:net}) to form the PU-Transformer body, where the channel dimension of each output follows: $32\rightarrow64\rightarrow128\rightarrow256\rightarrow256$. Particularly, in each Transformer Encoder, we only use the Positional Fusion block to encode the corresponding channel dimension (\ie, $C^{\prime}$ in Eq.~\ref{equ:g}), which remains the same in the subsequent operations. For all Positional Fusion blocks, the number of neighbors is empirically set to $k=20$ as used in previous works~\cite{wang2019dynamic,qian2021pu}.

In terms of the SC-MSA block, the primary way of choosing the shift-related parameters is inspired by the Non-local Network~\cite{wang2018non} and ECA-Net~\cite{Wang_2020_CVPR}. Specifically, a reduction ratio $\psi$~\cite{wang2018non} is introduced to generate the low-dimensional matrices in self-attention; following a similar method, the channel-wise width (\ie, channel dimension) of each split in SC-MSA is set as $w=C^\prime/\psi$. Moreover, since the channel dimension is usually set to a power of 2~\cite{Wang_2020_CVPR}, we simply set the channel-wise shift interval $d = w/2$. Therefore, the number of heads in SC-MSA becomes $M=2\psi-1$. In our implementation, $\psi=4$ is adopted in all SC-MSA blocks of PU-Transformer. 

\subsection{PU-Transformer Tail}
\label{sec:tail}
Based on the practical settings above, the input to the PU-Transformer tail (\ie, the output of the last Transformer Encoder) has a size of $N\times256$. Then, the periodic shuffling operation~\cite{shi2016real} reforms the channels and constructs a dense feature map of $rN\times256/r$, where $r$ is the upsampling scale. Finally, another MLP is applied to estimate the upsampled point cloud's 3D coordinates ($rN\times3$).


\section{Experiments}
\subsection{Settings}
\noindent \textbf{Training Details:}
In general, our PU-Transformer is implemented using Tensorflow~\cite{abadi2016tensorflow} with a single GeForce 2080 Ti GPU running on the Linux OS. In terms of the hyperparameters for training, we heavily adopt the settings from PU-GCN~\cite{qian2021pu} and Dis-PU~\cite{li2021point} for the experiments in Tab.~\ref{tab:pu1k} and Tab.~\ref{tab:pugan}, respectively. For example, we have a batch size of 64 for 100 training epochs, an initial learning rate of $1\times10^{-3}$ with a 0.7 decay rate, \etc~Moreover, we only use the modified Chamfer Distance loss~\cite{yifan2019patch} to train the PU-Transformer, minimizing the average closest point distance between the input set $\mathcal{P}\in\mathbb{R}^{N\times3}$ and the output set $\mathcal{S}\in\mathbb{R}^{rN\times3}$ for efficient and effective convergence. 

\noindent \textbf{Datasets:} 
Basically, we apply two 3D benchmarks for our experiments: 
\begin{itemize}
 \item \textbf{PU1K:} 
This is a new point cloud upsampling dataset introduced in PU-GCN~\cite{qian2021pu}. In general, the PU1K dataset incorporates 1,020 3D meshes for training and 127 3D meshes for testing, where most 3D meshes are collected from ShapeNetCore~\cite{chang2015shapenet} covering 50 object categories. To fit in with the patch-based upsampling pipeline~\cite{yifan2019patch}, the training data is generated from patches of 3D meshes via Poisson disk sampling. Specifically, the training data includes 69,000 samples, where each sample has 256 input points (low resolution) and a ground-truth of 1,024 points ($4\times$ high resolution).
 \item \textbf{PU-GAN Dataset:} 
This is an earlier dataset that was first used in PU-GAN~\cite{li2019pu} and generated in a similar way as PU1K but on a smaller scale. To be concrete, the training data comprises 24,000 samples (patches) collected from 120 3D meshes, while the testing data only contains 27 meshes. In addition to the PU1K dataset consisting of a large volume of data targeting the basic $4\times$ upsampling experiment, we conduct both $4\times$ and $16\times$ upsampling experiments based on the compact data of the PU-GAN dataset.

\end{itemize}

\noindent \textbf{Evaluation Metrics:}
As for the testing process, we follow common practice that has been utilized in previous point cloud upsampling works~\cite{yifan2019patch,li2019pu,li2021point,qian2021pu}. To be specific, at first, we cut the input point cloud into multiple seed patches covering all the $N$ points. Then, we apply the trained PU-Transformer model to upsample the seed patches with a scale of $r$. Finally, the farthest point sampling algorithm~\cite{qi2017pointnet} is used to combine all the upsampled patches as a dense output point cloud with $rN$ points. For the $4\times$ upsampling experiments in this paper, each testing sample has a low-resolution point cloud with 2,048 points, as well as a high-resolution one with 8,196 points. Coupled with the original 3D meshes, we quantitatively evaluate the upsampling performance of our PU-Transformer based on three widely used metrics: (i) Chamfer Distance (CD), (ii) Hausdorff Distance~\cite{berger2013benchmark} (HD), and (iii) Point-to-Surface Distance (P2F). A lower value under these metrics denotes better upsampling performance.


\begin{table}
\begin{center}
\captionsetup{font=small, skip=3pt}
\caption{Quantitative comparisons ($4\times$ Upsampling) to state-of-the-art methods on the \emph{PU1K} dataset~\cite{qian2021pu}. (\enquote{\textbf{CD}}: Chamfer Distance; \enquote{\textbf{HD}}: Hausdorff Distance; \enquote{\textbf{P2F}}: Point-to-Surface Distance. \enquote{\textbf{Model}}: model size; \enquote{\textbf{Time}}: average inference time per sample; \enquote{\textbf{Param.}}: number of parameters. $^\ast$: self-reproduced results, --: unknown data.)}
\resizebox{0.6\columnwidth}{!}{
\begin{tabular}{|c|ccc|ccc|}
\Xhline{3\arrayrulewidth}
\multirow{2}{*}{Methods} &\textbf{Model} &\textbf{Time}  &\textbf{Param.} &\multicolumn{3}{c|}{Results ($\times {10}^{-3}$)}  \\\cline{5-7}
&(MB) &($\times {10}^{-3}$s)  &($\times10^3$) &\textbf{CD} $\downarrow$ &\textbf{HD} $\downarrow$ &\textbf{P2F} $\downarrow$  \\\hline\xrowht{7pt}
PU-Net~\cite{yu2018pu} &10.1 &8.4  &812.0 &1.155 &15.170 &4.834                 \\
MPU~\cite{yifan2019patch} &6.2 &8.3  &76.2 &0.935 &13.327 &3.551                 \\
PU-GACNet~\cite{han2022pu} &-- &-- &\textbf{50.7} &0.665 &9.053 &2.429 \\
PU-GCN~\cite{qian2021pu} &\textbf{1.8} &\textbf{8.0}  &76.0 &0.585 &7.577 &2.499             \\
Dis-PU$^\ast$~\cite{li2021point} &13.2 &10.8  &1047.0 &0.485 &6.145 &1.802             \\\hline
\rowcolor{Gray} \textbf{Ours} &18.4 &9.9  &969.9 &\textbf{0.451} &\textbf{3.843} &\textbf{1.277}\\\Xhline{3\arrayrulewidth}      
\end{tabular}
\label{tab:pu1k}
}
\end{center}
\end{table}

\begin{table}
\begin{center}
\captionsetup{font=small, skip=3pt}
\caption{Quantitative comparisons to state-of-the-art methods on the \emph{PU-GAN} dataset~\cite{li2019pu}. (All metric units are ${10}^{-3}$. The best results are denoted in \textbf{bold}.)}
\resizebox{0.6\columnwidth}{!}{
\begin{tabular}{|c|ccc|ccc|}
\Xhline{3\arrayrulewidth}
\multirow{2}{*}{Methods} &\multicolumn{3}{c|}{$4\times$ Upsampling} &\multicolumn{3}{c|}{$16\times$ Upsampling}  \\\cline{2-7}
&\textbf{CD} $\downarrow$ &\textbf{HD} $\downarrow$ &\textbf{P2F} $\downarrow$ &\textbf{CD} $\downarrow$ &\textbf{HD} $\downarrow$ &\textbf{P2F} $\downarrow$ \\\hline
PU-Net~\cite{yu2018pu} &0.844 &7.061 &9.431 &0.699 &8.594 &11.619               \\
MPU~\cite{yifan2019patch} &0.632 &6.998 &6.199 &0.348 &7.187 &6.822                 \\
PU-GAN~\cite{li2019pu} &0.483 &5.323 &5.053 &0.269 &7.127 &6.306             \\
PU-GCN$^\ast$~\cite{qian2021pu} &0.357 &5.229 &3.628 &0.256 &5.938 &3.945\\
Dis-PU~\cite{li2021point} &0.315 &4.201 &4.149 &\textbf{0.199} &4.716 &4.249             \\\hline
\rowcolor{Gray}\textbf{Ours} &\textbf{0.273} &\textbf{2.605} &\textbf{1.836} &0.241 &\textbf{2.310} &\textbf{1.687}    \\\Xhline{3\arrayrulewidth}      
\end{tabular}
\label{tab:pugan}
}
\end{center}
\end{table}

\subsection{Point Cloud Upsampling Results}
\label{sec:results}
\noindent \textbf{PU1K:}
Table~\ref{tab:pu1k} shows the quantitative results of our PU-Transformer on the PU1K dataset. It can be seen that our approach outperforms other state-of-the-art methods on all three metrics. In terms of the Chamfer Distance metric, we achieve the best performance among all the tested networks, since the reported values of others are all higher than ours of 0.451. Under the other two metrics, the improvements of PU-Transformer are particularly significant: compared to the performance of the recent PU-GCN~\cite{qian2021pu}, our approach can almost \emph{halve} the values assessed under both the Hausdorff Distance (HD: 7.577$\rightarrow$3.843) and the Point-to-Surface Distance (P2F: 2.499$\rightarrow$1.277). 

\noindent \textbf{PU-GAN Dataset:}
We also conduct point cloud upsampling experiments using the dataset introduced in PU-GAN~\cite{li2019pu}. under more upsampling scales. As shown in Table~\ref{tab:pugan}, we achieve best performance under all three evaluation metrics for the $4\times$ upsampling experiment. However, in the $16\times$ upsampling test, we (CD: 0.241) are slightly behind the latest Dis-PU network~\cite{li2021point} (CD: 0.199) evaluated under the Chamfer Distance metric: the Dis-PU applies two CD-related items as its loss function, hence getting an edge for CD metric only. As for the results under Hausdorff Distance and Point-to-Surface Distance metrics, our PU-Transformer shows significant improvements again, where some values (\eg, P2F in $4\times$, HD and P2F in $16\times$) are even lower than \emph{half} of Dis-PU's results.

\begin{table}
\begin{center}
\captionsetup{font=small, skip=3pt}
\caption{Ablation study of the PU-Transformer's components tested on the \emph{PU1K} dataset~\cite{qian2021pu}. Specifically, models $A_1$-$A_3$ investigate the effects of the Positional Fusion block, models $B_1$-$B_3$ compare the results of different self-attention approaches, and models $C_1$-$C_3$ test the upsampling methods in the tail.}
\resizebox{0.8\columnwidth}{!}{
\begin{tabular}{c|c|c|c|ccc}
\Xhline{3\arrayrulewidth}
\multirow{2}{*}{models} & \multicolumn{2}{c|}{PU-Transformer Body} & \multirow{2}{*}{PU-Transformer Tail} & \multicolumn{3}{c}{Results ($\times {10}^{-3}$)} \\ \cline{2-3} \cline{5-7} 
& \multicolumn{1}{c|}{Positional Fusion} & Attention Type & & {\textbf{CD} $\downarrow$} &{\textbf{HD} $\downarrow$} & \textbf{P2F} $\downarrow$ \\ \hline
$A_1$ &None &SC-MSA & Shuffle &0.605 &6.477 &2.038\\
$A_2$ &$\mathcal{G}_{geo}$ &SC-MSA & Shuffle &0.558 &5.713 &1.751\\
$A_3$ &$\mathcal{G}_{feat}$ &SC-MSA & Shuffle &0.497 &4.164 &1.511\\\hline
$B_1$ &$\mathcal{G}_{geo}$ \& $\mathcal{G}_{feat}$ &SA~\cite{wang2018non} & Shuffle &0.526 &4.689 &1.492\\
$B_2$ &$\mathcal{G}_{geo}$ \& $\mathcal{G}_{feat}$ &OSA~\cite{guo2021pct} & Shuffle &0.509 &4.823 &1.586\\
$B_3$ &$\mathcal{G}_{geo}$ \& $\mathcal{G}_{feat}$ &MSA~\cite{vaswani2017attention} & Shuffle &0.498 &4.218 &1.427\\\hline
$C_1$ &$\mathcal{G}_{geo}$ \& $\mathcal{G}_{feat}$ &SC-MSA & MLPs~\cite{yu2018pu} &1.070 &8.732 &2.467\\
$C_2$ &$\mathcal{G}_{geo}$ \& $\mathcal{G}_{feat}$ &SC-MSA & DupGrid~\cite{yifan2019patch} &0.485 &3.966 &1.380\\
$C_3$ &$\mathcal{G}_{geo}$ \& $\mathcal{G}_{feat}$ &SC-MSA & NodeShuffle~\cite{qian2021pu}&0.505 &4.157 &1.404\\\hline
\rowcolor{Gray}\textbf{Full} &$\mathcal{G}_{geo}$ \& $\mathcal{G}_{feat}$ &SC-MSA &Shuffle &\textbf{0.451} &\textbf{3.843} &\textbf{1.277}\\\Xhline{3\arrayrulewidth}      
\end{tabular}
\label{tab:abl_parts}
}
\end{center}
\end{table}
\noindent \textbf{Overall Comparison:}
The experimental results in Table~\ref{tab:pu1k} and~\ref{tab:pugan} indicate the great effectiveness of our PU-Transformer. Moreover, given quantitative comparisons to CNN-based (\eg, GCN~\cite{li2019deepgcns}, GAN~\cite{goodfellow2014generative}) methods under different metrics, we demonstrate the superiority of transformers for point cloud upsampling by only exploiting the fine-grained feature representations of point cloud data.

\subsection{Ablation Studies}
\noindent \textbf{Effects of Components:}
Table~\ref{tab:abl_parts} shows the experiments that replace PU-Transformer's major components with different options. Specifically, we test three simplified models ($A_1$-$A_3$) regarding the Positional Encoding block output (Eq.~\ref{equ:g}), where employing both local \emph{geometric} $\mathcal{G}_{geo}$ and \emph{feature} $\mathcal{G}_{feat}$ context (model \enquote{Full}) provides better performance compared to the others. As for models $B_1$-$B_3$, we apply different self-attention approaches to the Transformer Encoder, where our proposed SC-MSA (Sec.~\ref{sec:body}) block shows higher effectiveness on point cloud upsampling. In terms of the upsampling method used in the PU-Transformer tail, some learning-based methods are evaluated as in models $C_1$-$C_3$. Particularly, with the help of our SC-MSA design, the simple yet efficient periodic shuffling operation (\ie, PixelShuffle~\cite{shi2016real}) indicates good effectiveness in obtaining a high-resolution feature map.   

\begin{table}[t]
\begin{center}
\captionsetup{font=small, skip=3pt}
\caption{The model's robustness to random noise tested on the \emph{PU1K} dataset~\cite{qian2021pu}, where the noise follows a normal distribution of $\mathcal{N}(0,1)$ and $\beta$ is the noise level.}
\resizebox{0.8\columnwidth}{!}{
\begin{tabular}{c|ccc|ccc|ccc}
\Xhline{3\arrayrulewidth}
\multirow{2}{*}{Methods} & \multicolumn{3}{c|}{$\beta=0.5\%$} & \multicolumn{3}{c|}{$\beta=1\%$} & \multicolumn{3}{c}{$\beta=2\%$} \\ \cline{2-10}
& {\textbf{CD} $\downarrow$} &{\textbf{HD} $\downarrow$} & \textbf{P2F} $\downarrow$ & {\textbf{CD} $\downarrow$} &{\textbf{HD} $\downarrow$} & \textbf{P2F} $\downarrow$ & {\textbf{CD} $\downarrow$} &{\textbf{HD} $\downarrow$} & \textbf{P2F} $\downarrow$ \\ \hline
PU-Net~\cite{yu2018pu} &1.006 &14.640 &5.253 &1.017 &14.998 &6.851 &1.333 &19.964 &10.378\\
MPU~\cite{yifan2019patch} &0.869 &12.524 &4.069 &0.907 &13.019 &5.625 &1.130 &16.252 &9.291\\
PU-GCN~\cite{qian2021pu} &0.621 &8.011 &3.524 &0.762 &9.553 &5.585 &1.107 &13.130 &9.378\\
Dis-PU~\cite{li2021point} &0.496 &6.268 &2.604 &\textbf{0.591} &7.944 &4.417 &\textbf{0.858} &10.960 &7.759\\\hline
\rowcolor{Gray}\textbf{Ours} &\textbf{0.453} &\textbf{4.052} &\textbf{2.127} &0.610 &\textbf{5.787} &\textbf{3.965} &1.058 &\textbf{9.948} &\textbf{7.551}\\
\Xhline{3\arrayrulewidth}      
\end{tabular}
\label{tab:abl_noise}
}
\end{center}
\end{table}

\begin{table}[t]
\begin{center}
\captionsetup{font=small, skip=3pt}
\caption{Model Complexity of PU-Transformer using different numbers of Transformer Encoders. (Tested on the \emph{PU1K} dataset~\cite{qian2021pu} with a single GeForce 2080 Ti GPU.)}
\resizebox{0.8\columnwidth}{!}{
\begin{tabular}{c|c|c|c|c|ccc}
\Xhline{3\arrayrulewidth}
\# Transformer &\multirow{2}{*}{\# Parameters} &\multirow{2}{*}{Model Size} & Training Speed & Inference Speed & \multicolumn{3}{c}{Results ($\times {10}^{-3}$)} \\ \cline{6-8}
Encoders&&&(per batch) & (per sample) & {\textbf{CD} $\downarrow$} &{\textbf{HD} $\downarrow$} & \textbf{P2F} $\downarrow$ \\ \hline
$L=3$ &438.3k &8.5M &12.2s &6.9ms &0.487 &4.081 &1.362\\
$L=4$ &547.3k &11.5M &15.9s &8.2ms &0.472 &4.010 &1.284\\
\rowcolor{Gray} $\bm{L=5}$ &969.9k &18.4M &23.5s &9.9ms &0.451 &\textbf{3.843} &1.277\\
$L=6$ &2634.4k &39.8M &40.3s &11.0ms &\textbf{0.434} &3.996 &\textbf{1.210}\\
\Xhline{3\arrayrulewidth}      
\end{tabular}
\label{tab:abl_complexity}
}
\end{center}
\end{table}
\noindent \textbf{Robustness to Noise:} As the PU-Transformer can upsample different types of point clouds, including real scanned data, it is necessary to verify our model's robustness to noise. Concretely, we test the pre-trained models by adding some random noise to the sparse input data, where the noise is generated from a standard normal distribution $\mathcal{N}(0,1)$ and multiplied with a factor $\beta$. In practice, we conduct the experiments under three noise levels: $\beta = 0.5\%$, $1\%$ and $2\%$. Table~\ref{tab:abl_noise} quantitatively compares the testing results of state-of-the-art methods. In most tested noise cases, our proposed PU-Transformer achieves the best performance, while Dis-PU~\cite{li2021point} shows robustness under the CD metric as explained in Sec.~\ref{sec:results}. 

\noindent \textbf{Model Complexity:}
Generally, our PU-Transformer is a light ($<$1M parameters) transformer model compared to image transformers~\cite{khan2021transformers,liu2021swin,dosovitskiy2020image} that usually have more than 50M parameters. In particular, we investigate the complexity of our PU-Transformer by utilizing different numbers of the Transformer Encoders. As shown in Table~\ref{tab:abl_complexity}, with more Transformer Encoders being applied, the model complexity increases rapidly, while the quantitative performance improves slowly. For a better balance between effectiveness and efficiency, we adopt the model with \emph{five} Transformer Encoders ($L=5$) in this work. Overall speaking, the PU-Transformer is a powerful and affordable transformer model for the point cloud upsampling task.   

\begin{figure}[t]
\begin{center}
\includegraphics[width=0.98\textwidth]{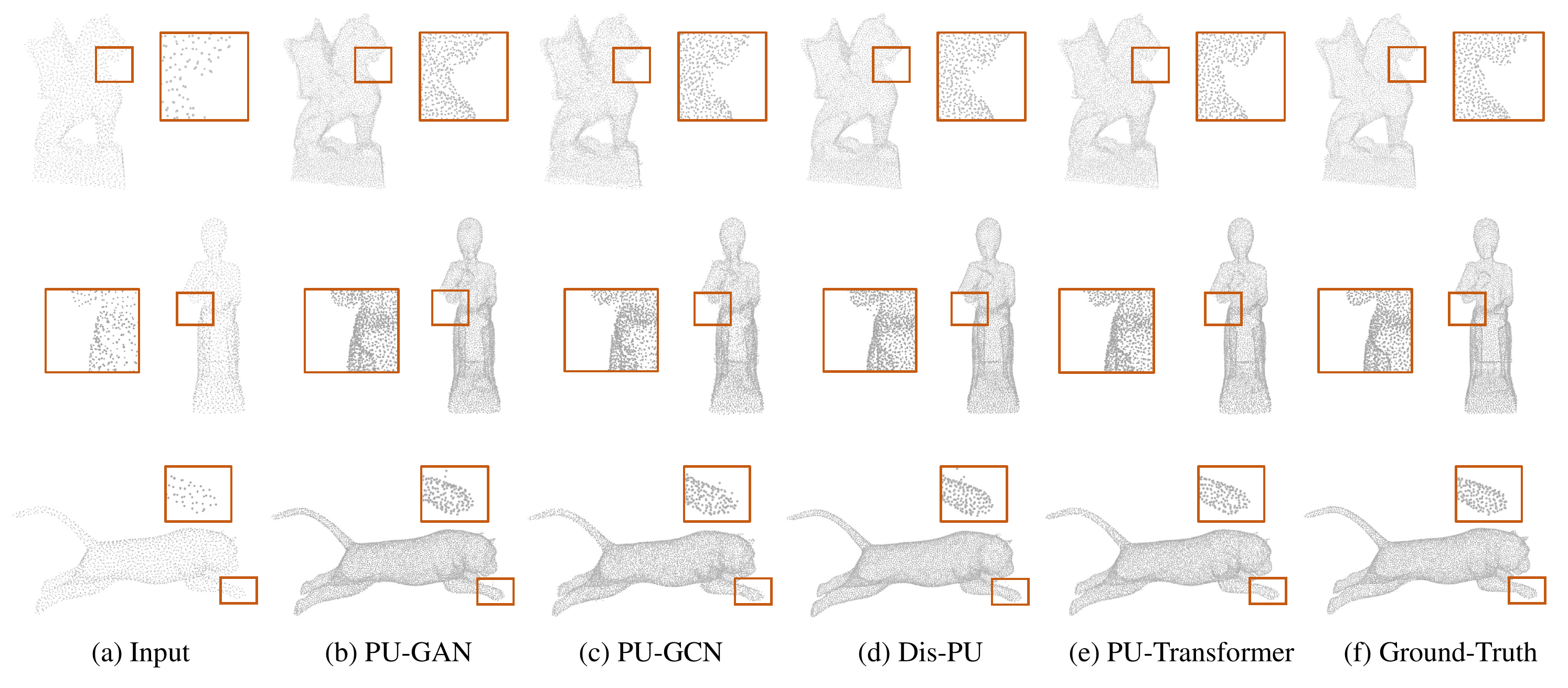}
\end{center}
\captionsetup{font=small}
\caption{Comparisons to state-of-the-art methods (PU-GAN~\cite{li2019pu}, PU-GCN~\cite{qian2021pu}, Dis-PU~\cite{li2021point}) in ($4\times$) upsampling \emph{synthetic} point cloud data using 2048 input points.}
\label{fig:vis}
\end{figure}

\begin{figure}[t]
\begin{center}
\includegraphics[width=\textwidth]{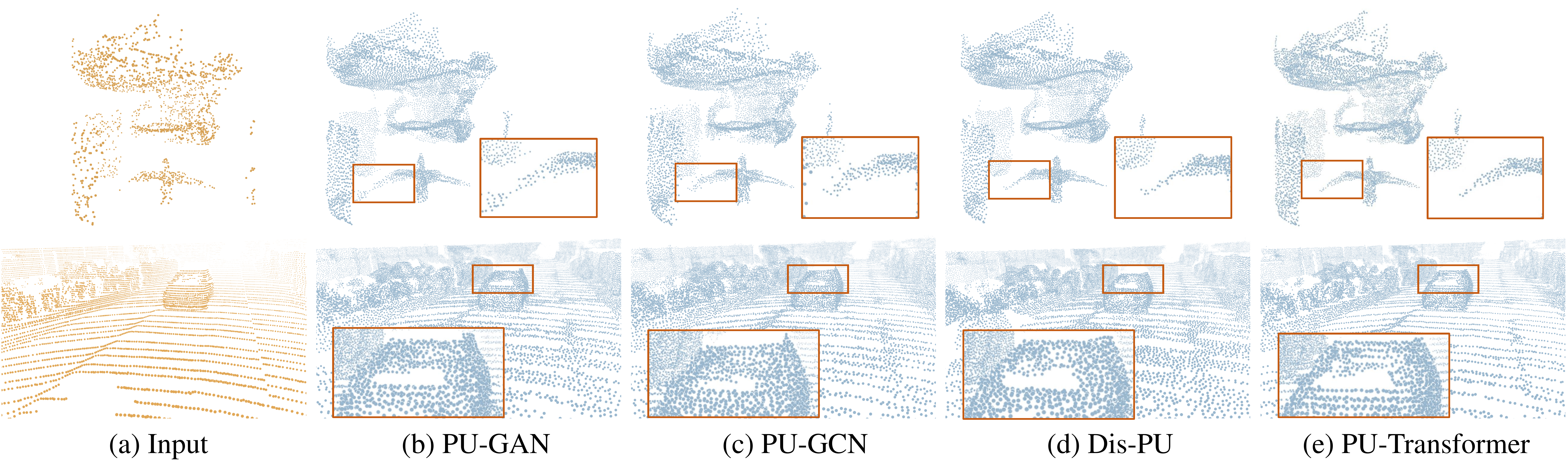}
\end{center}
\captionsetup{font=small}
\caption{Comparisons to state-of-the-art methods (PU-GAN~\cite{li2019pu}, PU-GCN~\cite{qian2021pu}, Dis-PU~\cite{li2021point}) in ($4\times$) upsampling \emph{real} point cloud data from ScanObjectNN~\cite{uy2019revisiting} dataset and SemanticKITTI~\cite{behley2019semantickitti} dataset.}
\label{fig:vis_scan}
\end{figure}

\subsection{Visualization}
\noindent \textbf{Qualitative Comparisons:}
The qualitative results of different point cloud upsampling models are presented in Fig.~\ref{fig:vis} and~\ref{fig:vis_scan}. Since we utilize the self-attention based structure to capture the point-wise dependencies from a global perspective, the PU-Transformer's output can better illustrate the overall contours of input point clouds producing fewer outliers (as shown in the zoom-in views of Fig.~\ref{fig:vis}). Particularly, based on the rich local context encoded by our Positional Fusion block, the PU-Transformer precisely upsamples the real point clouds (compared in Fig.~\ref{fig:vis_scan}), retaining a uniform distribution and much structural detail.

\noindent \textbf{Upsampling Different Input Sizes:}
Fig.~\ref{fig:res} shows the results of upsampling different sizes of point cloud data using PU-Transformer. Given a relatively low-resolution point cloud (\eg, 256 or 512 input points), our proposed model is still able to generate dense output with high-fidelity context (\eg, the head/foot of \enquote{Panda}). As the input size increases, the new points are uniformly distributed, covering the main flat areas (\eg, the body of \enquote{Panda}). 

\noindent \textbf{Upsampling Real Point Clouds:} In addition to Fig.~\ref{fig:vis_scan}, we provide more upsampling results ($4\times$ and $16\times$) on real point cloud samples (\ie, \enquote{chair}, \enquote{office}, \enquote{room}, \enquote{street}) from \emph{ScanObjectNN}~\cite{uy2019revisiting}, \emph{S3DIS}~\cite{armeni2017joint}, \emph{ScanNet}~\cite{dai2017scannet}, and \emph{SemanticKITTI}~\cite{behley2019semantickitti}, respectively. As Fig.~\ref{fig:scan} clearly illustrates, by addressing the sparsity and non-uniformity of raw inputs, not only is the overall quality of point clouds significantly improved, but also the representative features of object instances are enhanced. Particularly, the contours of upsampled object instances (\eg, \emph{tables} in \enquote{office/room}, \emph{cars} in \enquote{street}) are clearly distinct from the complex surroundings, obtaining high-fidelity details for visual analysis. More examples for visualization are included in the supplementary material.

\noindent
\begin{minipage}[t]{0.47\textwidth}
  \vspace{0pt}
  \centering
    \includegraphics[width=\columnwidth]{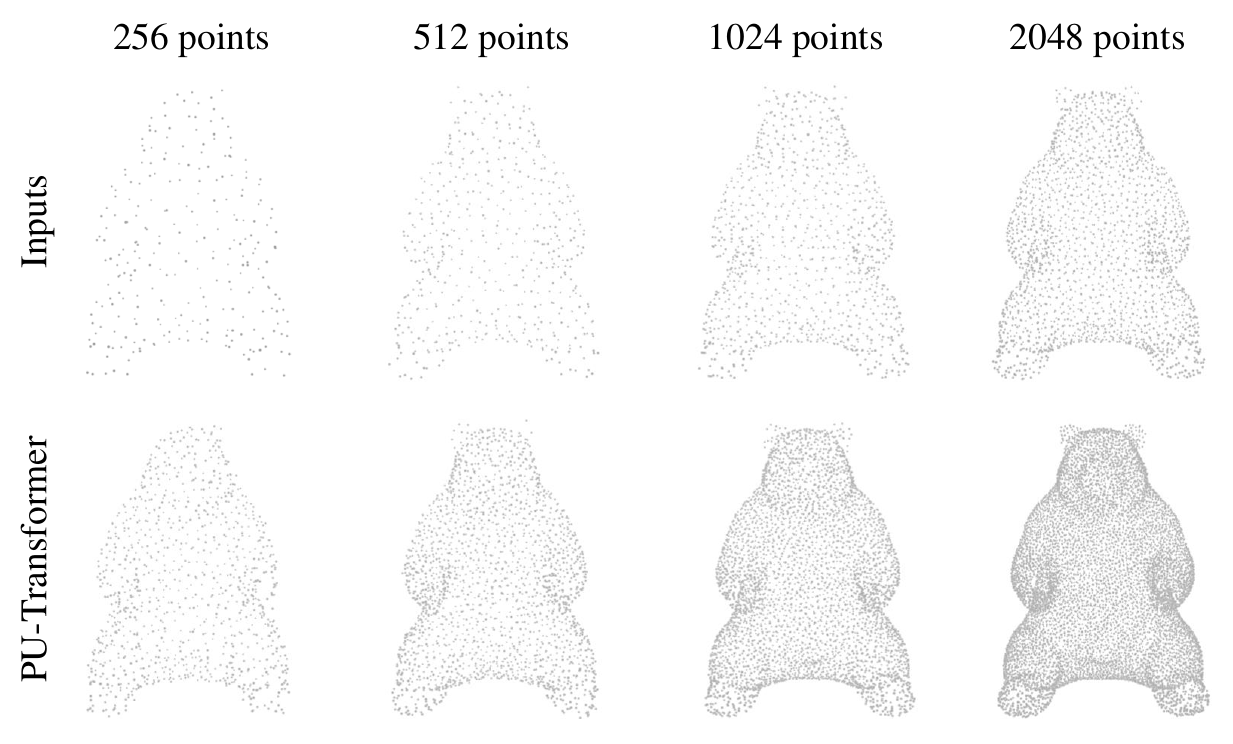}
    \captionof{figure}{\small PU-Transformer's $4\times$ upsampling results, given different sizes of input point cloud data.}
    \label{fig:res}
\end{minipage}%
\hfill%
\begin{minipage}[t]{0.47\textwidth}
  \vspace{0pt}
  \centering
    \includegraphics[width=0.9\columnwidth]{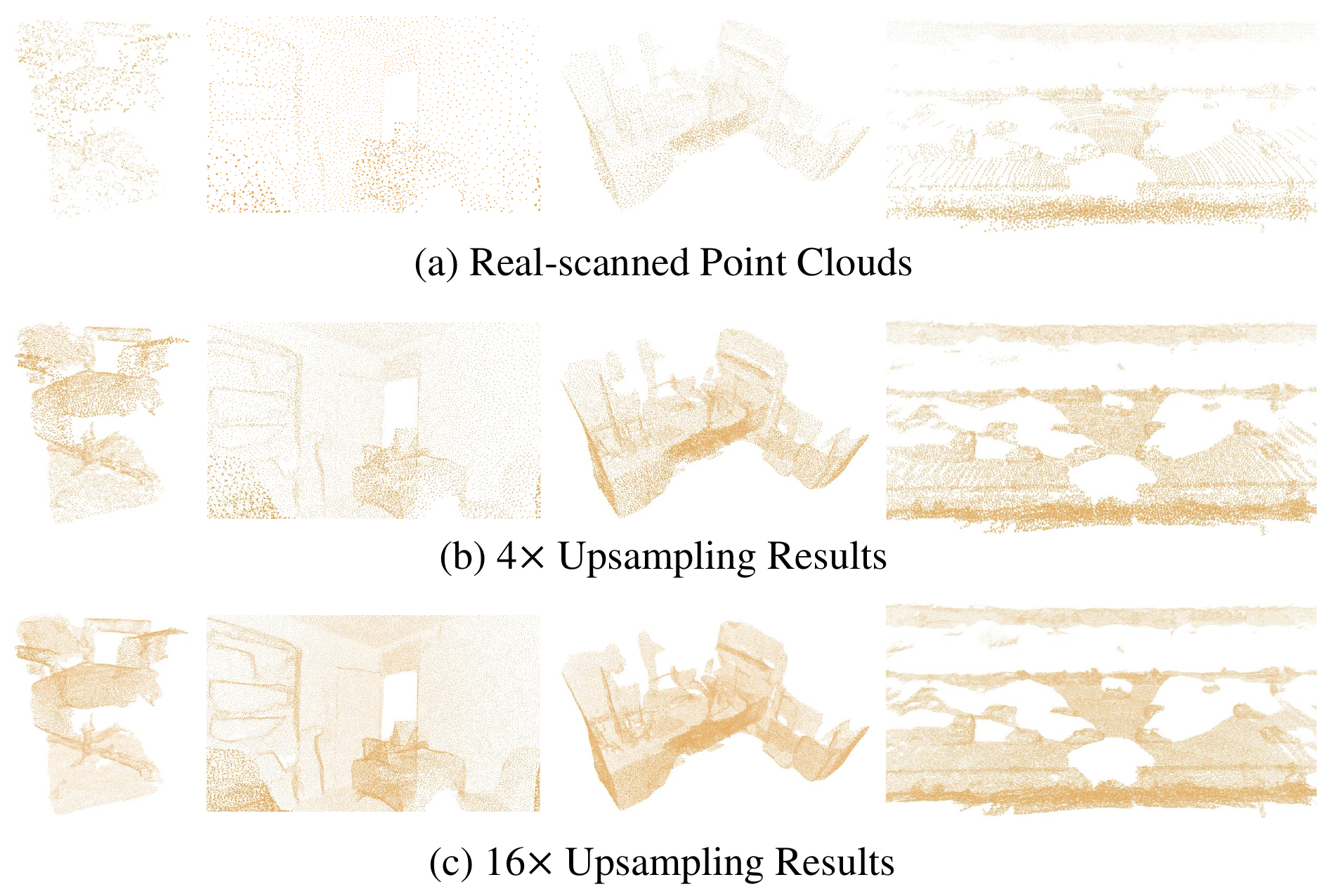}
    \captionof{figure}{\small PU-Transformer's $4\times$ and $16\times$ upsampling results, given different real point clouds.}
    \label{fig:scan}
\end{minipage}
\section{Limitations and Future Work}
\noindent 
\textbf{Upsampling Efficiency:}
Compared to the recent works such as Point Transformer~\cite{zhao2021point} ($\sim$7.76M parameters) or PoinTr~\cite{yu2021pointr} ($\sim$22.7M), PU-Transformer ($\sim$0.97M) is an efficient transformer for point clouds. However, it still consumes more parameters than some CNN-based counterparts~\cite{yu2018pu,qi2019deep,qian2021pu,li2021point} shown in Table~\ref{tab:pu1k}. As for inference speed, our approach is very close to others due to the succinct pipeline design, while methods that exploit complex network~\cite{li2019pu}, upsampling strategy~\cite{li2021point} or geometric calculations~\cite{qian2020pugeo} will be a bit slower.

\noindent
\textbf{Upsampling Flexibility:}
To generate different resolutions of output, our PU-Transformer may require some post-processing such as multiple inference iterations and farthest point sampling~\cite{qi2017pointnet}. For flexible point cloud upsampling, in future work, we will improve the adaptability of the PU-Transformer's body.

\noindent 
\textbf{Future Work:}
As a light-weight transformer targeting point clouds, our PU-Transformer has great potential in practice. For example, we could design a \emph{multi-functional} tail to solve different low-level vision problems such as upsampling, completion, and denoising. Moreover, we could further optimize the efficiency of the PU-Transformer in learning fine-grained point feature representations, benefiting the high-level visual analysis of large-scale point clouds.
\section{Conclusions}
\label{sec:concl}
This paper focuses on low-level vision for point cloud data in order to tackle its inherent \emph{sparsity} and \emph{irregularity}. Specifically, we propose a novel transformer-based model, PU-Transformer, targeting the fundamental point cloud upsampling task. Our PU-Transformer shows significant quantitative and qualitative improvements on different point cloud datasets compared to state-of-the-art CNN-based methods. By conducting related ablation studies and visualizations, we also analyze the effects and robustness of our approach. In the future, we expect to further optimize its efficiency for real-time applications and extend its adaptability in high-level 3D visual tasks.



\bibliographystyle{splncs}
\bibliography{egbib}

\end{document}